\documentclass{article}
\usepackage{spconf,amsmath,graphicx,url}
\usepackage{multirow}
\usepackage{booktabs}
\usepackage{algorithm}
\usepackage{algorithmicx}
\usepackage{algpseudocode} 
\usepackage{amsmath}

\title{Towards More Efficient and Effective Inference: \\The Joint Decision of Multi-Participants}
\name{Hui Zhu\textsuperscript{\textit{1,2}}, Zhulin An\textsuperscript{\textit{1,}}\sthanks{Corresponding author of this work.}, Kaiqiang Xu\textsuperscript{\textit{1}}, Xiaolong Hu\textsuperscript{\textit{1}}, Yongjun Xu\textsuperscript{\textit{1}}}

\address{\textsuperscript{1}Institute of Computing Technology, Chinese Academy of Sciences, Beijing, China\\\textsuperscript{2}University of Chinese Academy of Sciences, Beijing, China}
\begin{document}
%
\maketitle
\begin{abstract}
Existing approaches to improve the performances of convolutional neural networks by optimizing the local architectures or deepening the networks tend to increase the size of models significantly. In order to deploy and apply the neural networks to edge devices which are in great demand, reducing the scale of networks are quite crucial. However, It is easy to degrade the performance of image processing by compressing the networks. In this paper, we propose a method which is suitable for edge devices while improving the efficiency and effectiveness of inference. The joint decision of multi-participants, mainly contain multi-layers and multi-networks, can achieve higher classification accuracy (0.26\% on CIFAR-10 and 4.49\% on CIFAR-100 at most) with similar total number of parameters for classical convolutional neural networks. 
\end{abstract}
\begin{keywords}
Joint Decision, Multi-Layers, Multi-Networks, Image Classification
\end{keywords}

\section{Introduction}
Deep learning, which relies heavily on neural networks, has achieved huge success in the fields of computer vision and image processing. Recently, multitudinous neural architectures have been proposed to improve the performance of image processing. Bran-new architectures mainly come from human design such as VGG \cite{vgg}, ResNet \cite{resnet} and DenseNet \cite{densenet} or from automatic neural architecture search such as NASNet \cite{nasnet}, AmoebaNet \cite{AmoebaNet} and EfficientNet \cite{efficientnet}. With the improvement on the performance of image processing, the scale of neural networks is gradually increasing. 

Nevertheless, as an emerging field, edge computing tends to analyze and process data at the source rapidly to reduce costs and improve efficiency. Many researches attempt to reduce the scale of networks in order to deploy and apply them to edge devices such as wearable devices and IoT devices. The proposed methods are mainly limited to model compression \cite{Hansong2015,variational} and lightweight network design \cite{mobilenetv2,shufflenet}. These methods may have special requirements for the devices or only theoretically reduce the number of parameters and the amount of computations. Furthermore, these reductions in the scale of networks are often at the expense of the performance of image processing.

In this paper, we propose a new approach to address these problems, and thus, the results of image classification with higher accuracy can be achieved by using less number of parameters. We break up the large network into smaller parts and merge the information of more layers as multi-participants to make a joint decision. These small networks are more suitable for deploying to edge devices for the sake of limited storage and concurrent training. The experiment results also show that our approach is more efficient and effective for inference.

\begin{figure}
  \centering
  \includegraphics[width=1\linewidth]{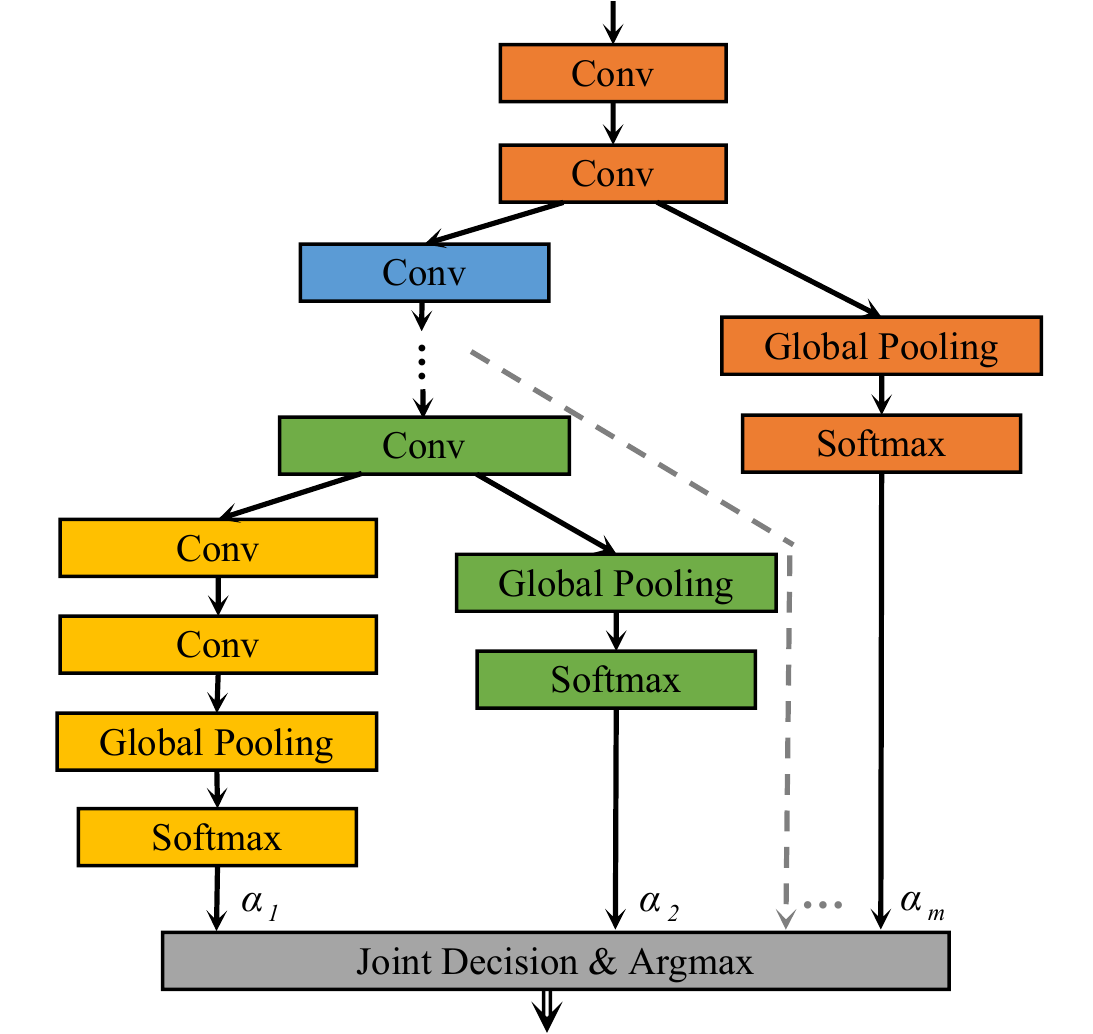}
  \caption{The joint decision of multi-layers. Different colors represent different scales of feature maps and $\alpha$ represents the weight of the layer in the final joint decision.}
\end{figure}

\begin{figure*}
  \centering
  \includegraphics[width=0.75\linewidth]{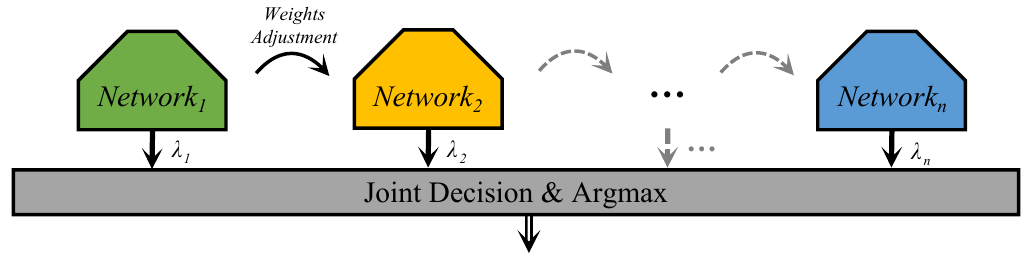}
  \caption{The joint decision of multi-networks. Different colors represent different Networks and $\lambda$ represents the weight of the network in the final joint decision. After the training for a network, we'll make small adjustments to the weights of the data.}
\end{figure*}

Our contributions are summarized as follows:
\begin{itemize}
\item We propose the joint decision of multi-participants which contain multi-layers and multi-networks. The effectiveness and robustness of a single network are improved, while the overall performance of multiple networks has better performance by this approach.
\item We propose clear rules for the architecture design and the training methods of every participant. 
\item Our method achieves better results for classification on CIFAR-10 and CIFAR-100 with a similar number of parameters and amount of computations.
\end{itemize}

\section{Related Work}
The methods to improve the performance of convolutional neural networks mainly focus on optimizing the convolutional neural architectures. Human-designed architectures such as short-cut connections \cite{resnet}, squeeze-excitation blocks \cite{senet} and automatic search algorithms such as ENAS \cite{enas}, DARTS \cite{darts} have been proposed to achieve some significant breakthroughs. At the same time, scaling down the models is also widely concerned to apply the networks on edge devices. To do this, many researches about model compression \cite{Hansong2015,variational} and lightweight network design \cite{mobilenetv2,shufflenet} are successively proposed. Our approach is inspired by Adaboost \cite{adaboost}, which trains several weak classifiers on the same training dataset and merge them to form the final classifier. It is more suitable for edge devices to achieve more efficient and effective inference and is orthogonal to these predecessor researches.

\section{Methods}
In this section, we propose our method, the joint decision of multi-participants which contain  
multi-layers and multi-networks. Furthermore, we propose the rules of the architecture design and the training methods for them.

\subsection{The Joint Decision of Multi-Layers}
Traditional convolutional neural networks extract features of images forward and finally achieve the image classification via the output of the last Softmax layer. Although the feature extraction tends to be more abstract by deeper layers in a network, we notice that the front layers can also solve some problems that deep ones cannot. We design the network in which multiple layers participate in the final decision together. 

The overview of joint decision of multi-layers is presented in Fig.1. We divide the network into several parts (suppose the number is $m$) according to the scales of the feature maps and they are represented by the different colors in the figure. Then we add the Global Average Pooling and the Softmax layer at the end of each part. In particular, a Batch-Norm layer and a activation function may be added simultaneously. This method will add few parameters and FLOPs, but gives voice for more layers in the network. 

Suppose that the weight of the original output of network is represented as $\alpha_{1}$. $k$ and $\mu$ represent the factors for regulation. Then the weight $\alpha_{m}$ of the output we added can be expressed as:
\begin{equation}
\alpha_{m} = \frac{\alpha_{1}}{k \cdot exp\left(m-1\right)}
\end{equation}

The loss function in the training can be expressed as:
\begin{equation}
Loss_{total} = \sum_{i=1}^{m} \alpha_{i} \cdot Loss_{i}
\end{equation}

And the final joint output can be expressed as:
\begin{equation}
Output_{total} = \alpha_{1} \cdot Output_{1} + \mu \sum_{i=2}^{m} \alpha_{i} \cdot Output_{i}
\end{equation}

\subsection{The Joint Decision of Multi-Networks}
It may cause a few misjudgments due to noise and other errors when the decision made by only one network. The joint decision made by multi-networks may effectively reduce these errors so as to improve the accuracy and robustness of the results. Based on this, we propose the method in which multiple networks participate in the final decision together.

The overview of joint decision of multi-networks is presented in Fig.2. Suppose that the number of total networks is $n$ and the accuracy for classification on training set of a network is $Acc$. Then, the weight $\lambda$ can be represented as:
\begin{equation}
\lambda_{n} = \frac{1}{\varepsilon}\ln\left( \frac{Acc_{n}}{1-Acc_{n}} \right)
\end{equation}

where $\varepsilon$ represents a factor to control $\lambda$ directly and further affect $W_{\xi}^{'}$. Specifically, for the first network, We train it normally on the original dataset and for the others, we will make small adjustments to the weights of data. We'll decrease the weights of correctly classified images and increase the weights of wrongly classified ones. In detail, we take an example of one image (its weight is represented as $W_{\xi}$) which is extracted from the total dataset. It initially has the same weight as the other images, but it may change depending on the classification results after the weight adjustment. The new weight $W_{\xi}^{'}$ for the $p$-th network can be represented as:
\begin{equation}
W_{\xi}^{'} =
\left\{  
	\begin{array}{lr}
	2exp\left(-\lambda_{p-1}\right)W_{\xi}  & \rm if \, right \\ 
	 \frac{1}{2}exp\left(\lambda_{p-1}\right)W_{\xi}  & \rm else
	\end{array}
\right. 
\end{equation}

When dealing with classification problems, The most common loss function is cross entropy:
\begin{equation}
Cross Entropy = -\sum_{t=1}^{N}y_{t}^{'}log\left(y_{t}\right)
\end{equation}

We can notice that it is easy to achieve the weight adjustment by modifying the value of labels.  The final joint output can be expressed as:
\begin{equation}
Output_{total} = \sum_{j=1}^{n} \lambda_{j} \cdot Output_{j}
\end{equation}

\subsection{The Joint Decision of Multi-Participants}
Similar to Adaboost, with the same number of parameters, we tend to use several small networks instead of one large network to make the final decision. We use $\gamma$ to record the total number of networks selected and the total number of parameters are similar to the original large network.

\begin{algorithm}[htb]
  \caption{The joint Decision of Multi-Participants}
  \begin{algorithmic}[1]
    \Require
    Number of Networks: $\mathcal \gamma$, Networks List: $N=\{\mathcal N_{1}$, $\mathcal N_{2}$, $\cdots$, $\mathcal N_{\gamma}\}$, Weight of Networks: $\mathcal \lambda$, Accuracy: $\mathcal Acc$, Weight of Data: $\mathcal W$, Number of data: $\mathcal T$.
    \Ensure  
   Final Decision: $\mathcal D_{fianl}$.
      \For {$i=1: \mathcal \gamma$}  
      	\State Construct the multi-layers network $\mathcal N_{i}$;
      	\State Calculate the joint accuracy of $\mathcal N_{i}$: $\mathcal Acc_{i}$;
      	\State Calculate the Weight of $\mathcal N_{i}$: $\mathcal \lambda_{i}$;
      		\For {$j=1: \mathcal T$}  
      	  		\State Update $\mathcal W_{\mathcal T}$;
      		\EndFor
      \EndFor 
      \State Calculate the joint output of $N$: $\mathcal D_{final}$;
   \State \Return{$\mathcal Acc_{D_{fianl}}$}
  \end{algorithmic}  
\end{algorithm}

The joint decision of multi-participants which contain multi-layers and multi-networks is described in Algorithm 1.

\begin{table}
  \caption{Comparison against the baselines of different networks on CIFAR-10. The performances for the joint decision of multi-layers with different $\mu$ are both improved.}\smallskip 
  \centering
  \resizebox{1\columnwidth}{!}{
  \begin{tabular}{l|c|c|c|c}
    \toprule
    \midrule  
    \multirow{2}{*}{\textbf{Networks}} & \textbf{Params} & \multicolumn{3}{c}{\textbf{Test Error} (\%)}\\
    \cmidrule{3-5}
    & (Mil.) & \textbf{Baseline}& \textbf{Ours ($\mu = 1$)}& \textbf{Ours ($\mu = 1/2$)}\\
	\midrule
	VGG-16 \cite{vgg} & 15.00 & 6.29$\pm$0.10 & 6.07$\pm$0.13 &   \textbf{6.05}$\pm$0.17  \\
    \midrule
    ResNet-18 \cite{resnet} & 11.18 & 4.00$\pm$0.11 & \textbf{3.87}$\pm$0.14 &  3.91$\pm$0.12 \\
    \midrule
    DenseNet-BC \cite{densenet} & 0.79 & 4.38$\pm$0.12 & 4.35$\pm$0.16 &  \textbf{4.26}$\pm$0.14 \\
    \midrule
    \bottomrule
  \end{tabular}
  }
\end{table}

\begin{table}
  \caption{Comparison against the results of original networks for ResNet-18 with different scaling factors on CIFAR-10. The performances for the joint decision of multi-networks with several appropriate $\gamma$ are improved.}\smallskip
  \centering
  \resizebox{1\columnwidth}{!}{
  \begin{tabular}{l|c|c|c|c}
    \toprule
    \midrule  
    \multirow{2}{*}{\textbf{Scaling Factor}} & \textbf{Params} & \textbf{Number of} & \textbf{Single}  &  \textbf{Total} \\
    &(Mil.) & \textbf{Networks $\gamma$} &  \textbf{Test Error} (\%) &  \textbf{Test Error} (\%)\\
	\midrule
    Original & 11.18 & 1 & 4.00$\pm$0.11 &  4.00$\pm$0.11 \\
    \midrule
    $\sqrt{1/2}$  & 5.78 & 2 & 4.22$\pm$0.09 &   \textbf{3.77}$\pm$0.16 \\
    \midrule
    \multirow{2}{*}{$1/2$} & \multirow{2}{*}{2.8} & 3 &  \multirow{2}{*}{4.90$\pm$0.24} &  \textbf{3.95}$\pm$0.08  \\
    \cmidrule{3-3} \cmidrule{5-5}
     & & 4 &  &   \textbf{3.79}$\pm$0.08  \\
    \midrule
    \multirow{2}{*}{$1/4$} & \multirow{2}{*}{0.7} & 5 & \multirow{2}{*}{6.61$\pm$0.21} &  5.24$\pm$0.06   \\
    \cmidrule{3-3} \cmidrule{5-5}
    & & 10 &  &  4.92$\pm$0.11   \\
    \midrule
    \bottomrule
  \end{tabular}
  }
\end{table}

\section{Experiments}
In this section, we introduce the implementation of experiments and report the performances of our methods. The experiments are mainly implemented in accordance with the methods mentioned in Sect.3 and we compare the networks designed by our methods with the original classical ones to prove the feasibility and effectiveness.

\subsection{Datasets}
We use CIFAR-10 and CIFAR-100 \cite{cifar} for image classification as basic datasets in our experiments. We normalize the images using channel means and standard deviations for preprocessing and apply a standard data augmentation scheme (zero-padded with 4 pixels on each side to obtain a $40\times40$ pixel image, then a $32\times32$ crop is randomly extracted and the image is randomly flipped horizontally). 

\begin{table*}
  \caption{Comparison against the results of classical convolutional neural networks on CIFAR-10 and CIFAR-100. The results about classification error on test datasets of the joint decision of multi-participants perform better while controlling the total number of parameters similar to the original networks.}\smallskip
  \centering
  \resizebox{2\columnwidth}{!}{
  \begin{tabular}{l||ccc||ccc}
    \toprule
    \midrule
    \multirow{3}{*}{\textbf{Networks}} & \multicolumn{3}{c||}{\textbf{Original Networks}} & \multicolumn{3}{c}{\textbf{The Joint Decision of Multi-Participants}} \\
    \cmidrule{2-7}
    & \textbf{Params} & \textbf{C10 Test Error} & \textbf{C100 Test Error} &  \textbf{Total Params} & \textbf{C10 Test Error} & \textbf{C100 Test Error} \\
     & (Mil.) & (\%) & (\%) & (Mil.)  & (\%) & (\%) \\
    \midrule
    VGG-16 \cite{vgg}  & 15.00  & 6.29$\pm$0.10 & 28.22$\pm$0.38  & 15.00 & 6.05$\pm$0.17 & 26.06$\pm$0.29 \\
    \midrule
    ResNet-18 \cite{resnet}  & 11.18  & 4.00$\pm$0.11 & 24.58$\pm$0.23  & 11.22 & 3.74$\pm$0.12 & 20.09$\pm$0.11 \\
    \midrule
    DenseNet-BC \cite{densenet}  & 0.79  & 4.38$\pm$0.12 & 21.96$\pm$0.19 &  0.73 & 4.14$\pm$0.13 & 21.77$\pm$0.19  \\
    \midrule
    MobileNetV2 \cite{mobilenetv2} & 2.29 & 6.32$\pm$0.19 & 25.68$\pm$0.23 & 2.30  & 6.16$\pm$0.18  & 23.45$\pm$0.21 \\
     \midrule
    \bottomrule
  \end{tabular}
  }
\end{table*}

\subsection{Training Methods}
Networks are trained on the full training dataset until convergence using Cutout \cite{cutout}. For a fair comparison, we retrain the original networks by the same training method with the networks designed by our methods. That is to say, all the networks which contain baselines and those modified by us are trained with a batch size of 128 using SGDR \cite{SGDR} with Nesterov's momentum for 511 epochs. The hyper-parameters of the methods are as follows: the cutout size is $16\times16$ for Cutout, $momentum=0.9$, $l_{max}=0.1$,$T_{0}=1$ and $T_{mult}=2$ for SGDR. We conducted every experiment more than three or four times and the mean classification error with standard deviation on the test dataset will be finally reported. 

\subsection{Crucial Preparations Before the Main Experiment}
In this subsection, we show some crucial preparations before the main experiment. We conducted confirmatory experiments separately according to the methods proposed above, and further determined the super parameters roughly.\smallskip

\noindent \textbf{The experiments for the joint decision of multi-layers.} We verified the performance of this method by several experiments for different classical networks on CIFAR-10. The networks contain VGG-16 (Reducing several FC layers), ResNet-18 and DenseNet-BC ($k=12,depth=100$). Together with the original networks, we got the results with two super parameters for $\mu$ ($\mu=1$ and $\mu=1/2$) and $k=1$. As is shown in Table 1, the performances of the networks with joint decision of multi-layers are better than the original ones. The difference for the performances of different $\mu$ is not obvious. Specifically, the increment of number of parameters by multi-layers can be ignored (about 0.01Mil.) so we didn't specify that in the table. By observing the results, we can definitely declare the effectiveness of our method.

\smallskip \noindent \textbf{The experiments for the joint decision of multi-networks.} We also verified the performance of this method by several experiments for ResNet-18 on CIFAR-10. We scaled the number of parameters of ResNet-18 to $1/2$, $1/4$, and $1/16$ of original by reducing the number of channels (the scaling factors in Table 2: $\sqrt{1/2}$, $1/2$ and $1/4$ of the original, respectively). Together with the original network, we got the results with super parameters for $\varepsilon=10$ and different $\gamma$ which should guarantee that the total number of parameters of the multi-networks is smaller or similar to the original network. As is shown in Table 2, the performance of the joint decision of multi-networks is better than the original with several appropriate $\gamma$. This leads to that the number of multiply networks should not be too many in the following main experiment.

\subsection{The Experiments for the Joint Decision of Multi-Participants and Results}
We conducted the experiments for the joint decision of multi-participants by combining both two parts mentioned above with the super parameters $\mu=1/2$ and $1\leq \gamma \leq 4$. We selected several classical convolutional neural networks to prove the excellence of our methods.

The comparison against the results of original classical convolutional neural networks on CIFAR-10 and CIFAR-100 is presented in Table 3. We show the comparison of the mean error with standard deviation on the test datasets while controlling the total number of parameters to be similar. Specifically, the number of parameters of the networks used to compare the test error on CIFAR-100 is slightly more than that on CIFAR-10, but we only show the one for CIFAR-10 in the table because the number of parameters are almost the same. 

We can clearly notice that the results of our method perform better. The accuracy can be improved by 0.26\% on CIFAR-10 and 4.49\% on CIFAR-100 at most with similar number of parameters (FLOPs is also similar because we only reduce the number of channels within the networks). 

What's more, the smaller single networks designed by our methods are more suitable for the edge devices. We no longer need to load the large networks at once for training or inference but are plagued by the limited storage. In addition, concurrent training and inference will make image processing more efficient and effective which we attribute to the joint decision of multi-participants.

\section{Conclusion}
We propose the joint decision of multi-participants, which mainly contain multi-layers and multi-networks. It is suitable for edge devices while improving the efficiency and effectiveness of inference. Our method can achieve higher classification accuracy with the similar number of parameters for classical convolutional neural networks and it is orthogonal to the predecessor researches.

\bibliographystyle{IEEEbib}
\bibliography{ref}

\begin{thebibliography}{10}

\bibitem{vgg}
Karen Simonyan and Andrew Zisserman,
\newblock ``Very deep convolutional networks for large-scale image
  recognition,''
\newblock in {\em 3rd International Conference on Learning Representations,
  {ICLR} 2015, San Diego, CA, USA, May 7-9, 2015, Conference Track
  Proceedings}, 2015.

\bibitem{resnet}
Kaiming He, Xiangyu Zhang, Shaoqing Ren, and Jian Sun,
\newblock ``Deep residual learning for image recognition,''
\newblock in {\em 2016 {IEEE} Conference on Computer Vision and Pattern
  Recognition, {CVPR} 2016, Las Vegas, NV, USA, June 27-30, 2016}, 2016, pp.
  770--778.

\bibitem{densenet}
Gao Huang, Zhuang Liu, Laurens van~der Maaten, and Kilian~Q. Weinberger,
\newblock ``Densely connected convolutional networks,''
\newblock in {\em 2017 {IEEE} Conference on Computer Vision and Pattern
  Recognition, {CVPR} 2017, Honolulu, HI, USA, July 21-26, 2017}, 2017, pp.
  2261--2269.

\bibitem{nasnet}
Barret Zoph, Vijay Vasudevan, Jonathon Shlens, and Quoc~V. Le,
\newblock ``Learning transferable architectures for scalable image
  recognition,''
\newblock in {\em 2018 {IEEE} Conference on Computer Vision and Pattern
  Recognition, {CVPR} 2018, Salt Lake City, UT, USA, June 18-22, 2018}, 2018,
  pp. 8697--8710.

\bibitem{AmoebaNet}
Esteban Real, Alok Aggarwal, Yanping Huang, and Quoc~V. Le,
\newblock ``Regularized evolution for image classifier architecture search,''
\newblock {\em CoRR}, vol. abs/1802.01548, 2018.

\bibitem{efficientnet}
Mingxing Tan and Quoc~V. Le,
\newblock ``Efficientnet: Rethinking model scaling for convolutional neural
  networks,''
\newblock in {\em Proceedings of the 36th International Conference on Machine
  Learning, {ICML} 2019, 9-15 June 2019, Long Beach, California, {USA}}, 2019,
  pp. 6105--6114.

\bibitem{Hansong2015}
Song Han, Jeff Pool, John Tran, and William~J. Dally,
\newblock ``Learning both weights and connections for efficient neural
  network,''
\newblock in {\em Advances in Neural Information Processing Systems 28: Annual
  Conference on Neural Information Processing Systems 2015, December 7-12,
  2015, Montreal, Quebec, Canada}, 2015, pp. 1135--1143.

\bibitem{variational}
Chenglong Zhao, Bingbing Ni, Jian Zhang, Qiwei Zhao, Wenjun Zhang, and Qi~Tian,
\newblock ``Variational convolutional neural network pruning,''
\newblock in {\em {IEEE} Conference on Computer Vision and Pattern Recognition,
  {CVPR} 2019, Long Beach, CA, USA, June 16-20, 2019}, 2019, pp. 2780--2789.

\bibitem{mobilenetv2}
Mark Sandler, Andrew~G. Howard, Menglong Zhu, Andrey Zhmoginov, and
  Liang{-}Chieh Chen,
\newblock ``Mobilenetv2: Inverted residuals and linear bottlenecks,''
\newblock in {\em 2018 {IEEE} Conference on Computer Vision and Pattern
  Recognition, {CVPR} 2018, Salt Lake City, UT, USA, June 18-22, 2018}, 2018,
  pp. 4510--4520.

\bibitem{shufflenet}
Xiangyu Zhang, Xinyu Zhou, Mengxiao Lin, and Jian Sun,
\newblock ``Shufflenet: An extremely efficient convolutional neural network for
  mobile devices,''
\newblock in {\em 2018 {IEEE} Conference on Computer Vision and Pattern
  Recognition, {CVPR} 2018, Salt Lake City, UT, USA, June 18-22, 2018}, 2018,
  pp. 6848--6856.

\bibitem{senet}
Jie Hu, Li~Shen, and Gang Sun,
\newblock ``Squeeze-and-excitation networks,''
\newblock in {\em 2018 {IEEE} Conference on Computer Vision and Pattern
  Recognition, {CVPR} 2018, Salt Lake City, UT, USA, June 18-22, 2018}, 2018,
  pp. 7132--7141.

\bibitem{enas}
Hieu Pham, Melody~Y. Guan, Barret Zoph, Quoc~V. Le, and Jeff Dean,
\newblock ``Efficient neural architecture search via parameter sharing,''
\newblock in {\em Proceedings of the 35th International Conference on Machine
  Learning, {ICML} 2018, Stockholmsm{\"{a}}ssan, Stockholm, Sweden, July 10-15,
  2018}, 2018, pp. 4092--4101.

\bibitem{darts}
Hanxiao Liu, Karen Simonyan, and Yiming Yang,
\newblock ``{DARTS:} differentiable architecture search,''
\newblock {\em CoRR}, vol. abs/1806.09055, 2018.

\bibitem{adaboost}
Yoav Freund and Robert~E. Schapire,
\newblock ``A decision-theoretic generalization of on-line learning and an
  application to boosting,''
\newblock in {\em Computational Learning Theory, Second European Conference,
  EuroCOLT '95, Barcelona, Spain, March 13-15, 1995, Proceedings}, 1995, pp.
  23--37.

\bibitem{cifar}
A.~Krizhevsky and G.~Hinton,
\newblock ``Learning multiple layers of features from tiny images,''
\newblock {\em Computer Science Department, University of Toronto, Tech. Rep},
  vol. 1, 01 2009.

\bibitem{cutout}
Terrance Devries and Graham~W. Taylor,
\newblock ``Improved regularization of convolutional neural networks with
  cutout,''
\newblock {\em CoRR}, vol. abs/1708.04552, 2017.

\bibitem{SGDR}
Ilya Loshchilov and Frank Hutter,
\newblock ``{SGDR:} stochastic gradient descent with warm restarts,''
\newblock in {\em 5th International Conference on Learning Representations,
  {ICLR} 2017, Toulon, France, April 24-26, 2017, Conference Track
  Proceedings}, 2017.

\end{thebibliography}

\end{document}